\documentclass[11pt]{article}
\usepackage{makecell} 

\providecommand{\keywords}[1]{%
  \par\vspace{0.5em}%
  \noindent\textbf{Keywords: } #1\par
}

\usepackage[authoryear,round]{natbib}
\usepackage{graphicx}
\usepackage{booktabs}
\usepackage{makecell}
\usepackage{hyperref}
\usepackage{amsmath, amssymb}

\title{Enhancing Clinical Note Generation with ICD-10, Clinical Ontology Knowledge Graphs, and Chain-of-Thought Prompting Using GPT-4} 

\author{Ivan Makohon,$^{1}$ Mohamad Najafi,$^{1}$ Jian Wu,$^{1}$\\
Mathias Brochhausen,$^{2}$ Yaohang Li$^{1\ast}$\\
{$^{1}$Computer Science, Old Dominion University}\\
{4700 Elkhorn Ave, Norfolk, VA 23529, USA}\\
{$^{2}$Biomedical Informatics, University of Arkansas for Medical Sciences}\\
{4301 W. Markham St. Little Rock, AR 72205, USA}\\
{$^\ast$To whom correspondence should be addressed;}\\
{E-mail:  yaohang@cs.odu.edu}
}
\date{}

\begin{document} 

\maketitle


\begin{abstract}
In the past decade a surge in the amount of electronic health record (EHR) data in the United States, attributed to a favorable policy environment created by the Health Information Technology for Economic and Clinical Health (HITECH) Act of 2009 and the 21st Century Cures Act of 2016.  
Clinical notes for patients’ assessments, diagnoses, and treatments are captured in these EHRs in free-form text by physicians, who spend a considerable amount of time entering and editing them. 
Manually writing clinical notes takes a considerable amount of a doctor's valuable time, increasing the patient’s waiting time and possibly delaying diagnoses.  
Large language models (LLMs) possess the ability to generate news articles that closely resemble human-written ones.  
We investigate the usage of Chain-of-Thought (CoT) prompt engineering to improve the LLM’s response in clinical note generation.
In our prompts, we use as input International Classification of Diseases (ICD) codes and basic patient information. We investigate a strategy that combines the traditional CoT with semantic search results to improve the quality of generated clinical notes.
Additionally, we infuse a knowledge graph (KG) built from clinical ontology to further enrich the domain-specific knowledge of generated clinical notes.
We test our prompting technique on six clinical cases from the CodiEsp test dataset using GPT-4 and our results show that it outperformed the clinical notes generated by standard one-shot prompts. 
\end{abstract}

\section{Introduction}

In the past decade, there has been a surge in the amount of electronic health record (EHR) data in the United States.  
In 2008, only 42\% of physicians had access to an EHR~\citep{onc_ehr_adoption}.  This figure has now risen to 88\% as reported in 2021~\citep{onc_ehr_adoption}.  
This increase may be attributed to a favorable policy environment created by the Health Information Technology for Economic and Clinical Health (HITECH) Act of 2009~\citep{burde2011hitech} and the 21st Century Cures Act of 2016~\citep{fda_cures_act}. 

Clinical notes for patients’ assessments, diagnoses, and treatments are captured in these EHRs in free-form text by physicians, who spend a considerable amount of time entering them into computers.  
These notes offer valuable insights based on real-time observed data, which have shown to enhance the predictive capabilities of medical decision-making models~\citep{hlthaff.2018.0728}.  

Despite the rich information contained in these notes, it is likely some details are excluded from publicly available LLMs due to restrictions on access to their content, a consequence of the Health Insurance Portability and Accountability Act (HIPAA) of 1996~\citep{moore2019hipaa}. 
HIPAA plays a crucial role in safeguarding the privacy and security of patients’ protected health information (PHI) in the context of clinical notes. One key outcome of HIPAA is the standardization of clinical text de-identification, which allows sensitive patient information to be stripped from records so the data can be safely reused. This de-identification process itself is a complex natural language processing (NLP) task, often requiring substantial time~\citep{Negash2023DeidentificationOF}. As a result, datasets used for pretraining LLMs are typically assumed to exclude PHI.

Despite these technological and regulatory advancements, the effectiveness of clinical note documentation has not kept pace~\citep{arndt2017ehr, gellert2015scribes, kroth2019burnout}. Among the factors contributing to this stagnation is physician burnout, a growing concern that impacts both provider well-being and patient care. Physicians often experience emotional exhaustion, reduced motivation, and a sense of detachment from their patients, largely due to the demanding and high-pressure nature of their work. One significant driver of this burnout is the administrative burden associated with EHRs, particularly the excessive time spent on data entry and clinical documentation~\citep{arndt2017ehr, gellert2015scribes, kroth2019burnout, liu2018notes}.


The medical scribe industry has emerged to handle the burdensome documentation tasks behind the scenes~\citep{gellert2015scribes}. However, relying on non-professional scribes presents challenges, as these unlicensed individuals, working under clinician supervision, may lack the medical expertise necessary for accurate EHR documentation, reflecting broader gaps in understanding of their impact and effectiveness~\citep{14302}. To address the physician burnout challenge, we focus our attention on LLMs given remarkable progress has been made in recent years with some observations suggesting that they exhibit more powerful reasoning abilities as the model size increases~\citep{brown2020fewshot}. Clinicians show growing interest in adopting LLMs for medical question answering, clinical note summarization, diagnostic assistance, and clinical note documentation, appreciating their potential to reduce documentation burden and improve efficiency, as highlighted in recent reviews~\citep{bioengineering12060631, 2ed5b7d99b5c4cf6a62adc81a1648a32, healthcare13060603}.

In this paper, we investigate the use of large language models (LLMs) to generate medical notes.
We begin by leveraging International Classification of Diseases (ICD) codes as initial prompts for the LLM.
To enhance the quality and relevance of the generated notes, we apply Chain-of-Thought (CoT) prompting, using example clinical cases to guide GPT-4’s reasoning process.
In addition, we explore the integration of SNOMED CT OWL expressions as a clinical ontology knowledge graph (KG) prompt.
The KG offers a structured, machine-readable representation of medical concepts and their interrelationships, such as disease hierarchies, symptom-diagnosis associations, treatment pathways, and comorbid conditions.
We evaluated the performance of GPT-4 in generating the patient’s current history of present illness (HPI) based on specific task instructions, using diagnosis codes and relevant patient information as input.
To benchmark our approach, we apply the CoT prompting technique to six clinical cases from the CodiEsp test dataset.


\section{Related Works}
The rapid advancements in LLMs have greatly enhanced their ability to comprehend patterns and relationships between words and phrases more effectively by developing a general understanding of grammar, syntax, and semantic relationships to generate text, bringing their output closer to human-level quality in areas of news compositions, story generation, and code generation ~\citep{wu2023survey}.  
LLMs like GPT-3 ~\citep{brown2020fewshot} and GPT-4 ~\citep{openai2023gpt4} have demonstrated impressive performance on downstream NLP tasks, even in zero-shot and few-shot settings.  
With its substantial capacity, it possesses the ability to generate news articles that closely resemble human-written ones, making it difficult to distinguish between the two ~\citep{brown2020fewshot}.  
This poses a particular challenge in detecting LLM-generated text, which is crucial for ensuring responsible AI governance ~\citep{wu2023survey}.  
GPT-4 is said to adhere more closely to guardrails, ensuring a higher level of responsible text generation. 

Prompt engineering (or In-Context Prompting) ~\citep{ye2023investigatingeffectivenesstaskagnosticprefix, wei2022cot} emerged as a recent field focused on crafting and refining prompts to effectively harness techniques aimed at interacting with LLM to guide its behavior towards specific goals, without making changes to the model weights. 
Since the recent releases of LLMs, Google researchers recently revolutionized a prompting strategy in solving word problems across five different LLMs ~\citep{zhang2023multimodal}. Several prompt engineering techniques ~\citep{li2022survey, sahoo2024prompt} have emerged and significantly improves the performance of LLMs on many natural language generation tasks.  
Recent studies, such as CoT ~\citep{wei2022cot, zhang2023multimodal}, Tree-of-Thought (ToT) ~\citep{yao2023tree, long2023tree}, and Graph-of-Thought (GoT) ~\citep{besta2024graph, yao2023graph} have shown to improve the reasoning and accuracy performance of LLMs by providing rationales for a given word or phrase ~\citep{wei2022cot, zhang2023multimodal, yao2023tree, long2023tree, besta2024graph, yao2023graph}.  
Although self-verification ~\citep{weng2023selfverification} and self-consistency ~\citep{wang2022selfconsistency} have enhanced performance in CoT prompting, recent prompting techniques such as ToT ~\citep{yao2023tree, long2023tree} and GoT ~\citep{besta2024graph, yao2023graph} have shown improvement, though their effectiveness is still being assessed.  CoT ~\citep{lievin2022medical} has demonstrated that LLMs are capable of reasoning through multiple-choice questions on medical board exams.  
For our purposes, can it reason about ICD codes along with some patient information to generate clinical notes?

Previous endeavors have demonstrated that employing an attention mechanism in a multi-label classification task can effectively yield ICD codes from clinical notes ~\citep{makohon2021multi} and shows that numerous prior research endeavors have revolved around classifying ICD codes using clinical notes as their primary input data~\citep{LiICD9, NIU2023104396, Wu9837396}. 
Our work in this paper is to reverse this process by generating comprehensive clinical notes, guided by provided ICD codes and supplemented basic patient information using instructional prompting techniques. 
In a recent study ~\citep{lee2024medicalcodes}, LLMs were investigated using zero-shot prompting to predict ICD-10 codes. 
ICD-10 codes were provided in their prompt: ``Predict these ICD-10 codes to the best of your ability” without any patient information or a clear instruction task to generate clinical notes. 
Based on their outputs, the LLMs outputs predicted just the ICD codes titles, not actual patient clinical notes.

Additionally, recent studies have explored the use of LLMs for generating clinical notes with the use of prompt engineering. 
These include leveraging LLMs to convert transcribed interactions into structured notes through structured prompting and integration of supplementary data for improved quality ~\citep{biswas2024clinicaldoc}, developing a specialized medical LLM to understand and summarize medical conversations using zero-shot prompt for note generation ~\citep{yuan2024mednote}, and providing rapid access to medical information via a chatbot that utilizes a predefined system prompt to perform contextual searches and Retrieval-Augmented Generation (RAG) techniques ~\citep{leong2024genai}.  
Some of the challenges in clinical note generation through use of LLMs are captured in this study ~\citep{wang2024rlnotes}, which highlights the feasibility of training efficient open-source LLMs for clinical note generation, with opportunities for further exploration in domain adaptation, data selection, and reinforcement learning from human feedback (RLHF).  
RLHF helps align LLMs with human preferences and can be applied in two ways: outcome-supervised, which focuses on improving the overall quality of the text, and process-supervised, which provides more detailed guidance on specific text components, such as reasoning steps, as seen in approaches like InstructGPT ~\citep{li2022survey}.  

We conduct experiments on the closed-source GPT-4 using semantic searches and the CoT prompting technique to query similar clinical cases based on the given ICD codes or text references.  
We further explore the integration of clinical ontology KG into the prompts to enrich the LLM input with well-defined medical relationships and constraints.
To the best of our knowledge, we are the first to perform experiments of this kind using diagnosis codes (ICD codes) as input along with basic patient information to generate clinical notes using LLMs and CoT prompting instructions.  
We seek to answer our research question: Can LLMs reason about ICD codes to guide the generation of clinical notes using instruction prompting?

\section{Methodology}
This paper explores a method for guiding the generation of clinical notes using an LLM while providing patient information, ICD codes, and task instructions as input, medical relationships between ICD codes and constraints derived from a KG, utilizing CoT with examples of clinical notes diagnosis returned by semantic search.

\subsection{Semantic Search \& Clinical Cases}
CodiEsp, introduced during the CodiEsp track for CLEF eHealth 2020, is recognized as a gold-standard annotated data source ~\citep{miranda2020codiesp}.  
The dataset is comprised of 1,000 clinical cases, where the clinical notes are translated from Spanish to English.  
It encompasses both ICD-10 CM and PCS codes, distributed across three randomly sampled datasets: the Training set contains 500 clinical cases.  
The Development (validation) and the Test set each contains 250 clinical cases.  
The text-reference column consists of text used during the annotated process using the Brat visualization tool ~\citep{stenetorp2012brat}.  
Hereafter, we will refer to text-reference column as the Text Reference.

\begin{figure}[ht]
\centering
\includegraphics[width=\linewidth]{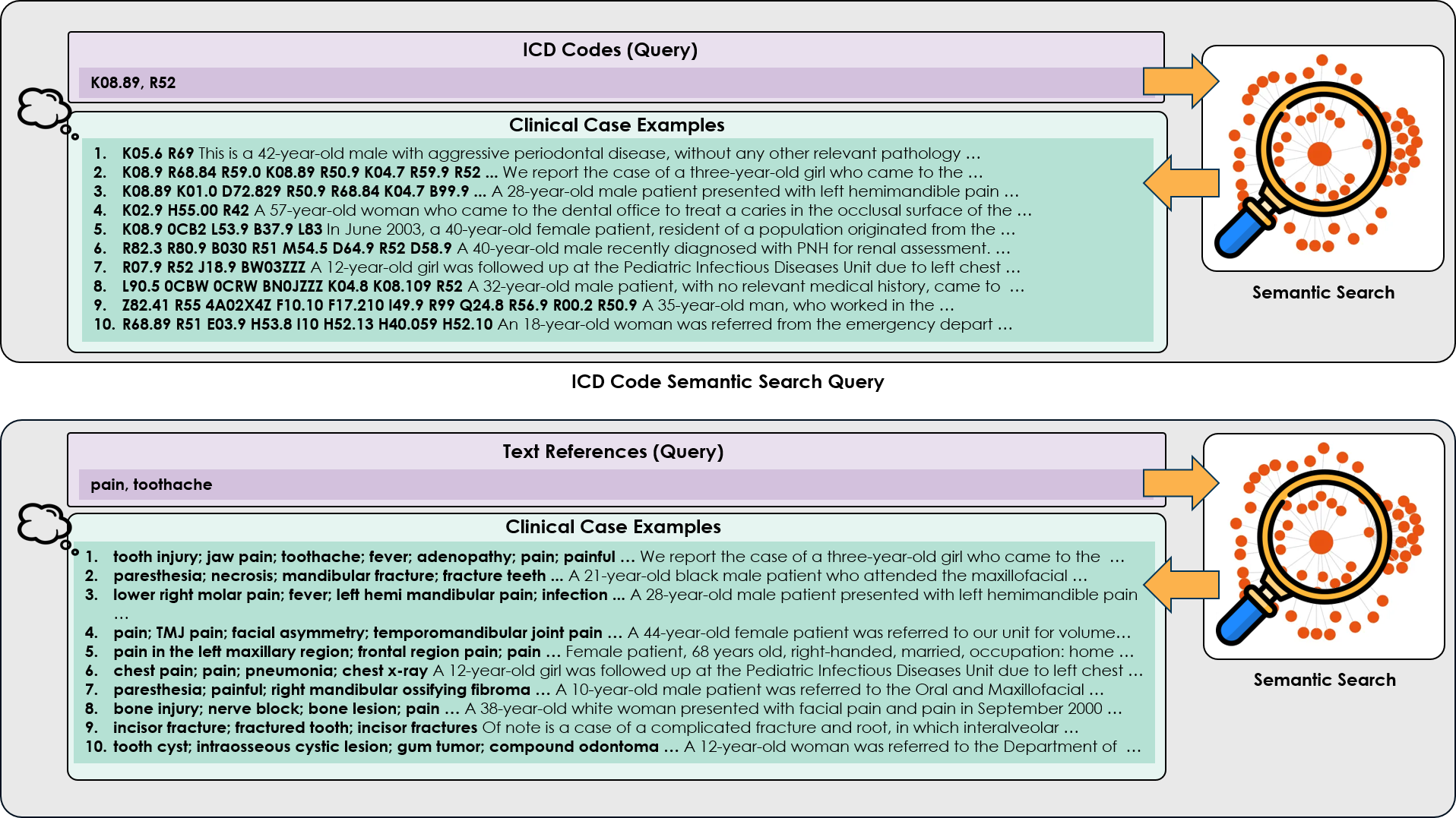}
\caption{An illustration of the Semantic Search query used to search for similarities within the embedding space.}
\label{fig:fig1}
\end{figure}

We combine CodiEsp’s training and validation datasets (750) for our semantic search embedding query, while reserving the test dataset (250) for selecting six clinical cases samples for evaluating ground-truth against generative text.  
We converted the texts in the combined dataset of 750 clinical cases into numerical vector representations with OpenAI’s text embedding model (text-embedding-3-small).  
Our objective is to leverage the embedding-driven retrieval to tap into the rich semantic features present in other clinical cases.  
This is achieved through the use of query ICD codes or text references, which facilitates the provision of clinical case examples for use as “thoughts” in our CoT prompt. 
As we will demonstrate, these semantic searches provide an efficient approach to identifying examples resembling the examples in the prompts.
The embedded query (ICD code or text reference) is used to pinpoint the most relevant clinical cases by assessing their proximity within the embedding space, utilizing document similarity to rank and present the top-n most suitable clinical cases.
For each query, the cosine similarity is used to identify the top-$n$ most similar clinical cases. 
Figure~\ref{fig:fig1} shows an illustration of the Semantic Search query used to search for similarities within the embedding space. The top illustration shows the ICD Code query. The bottom illustration shows the Text References query. The violet highlight depicts the type of inputs for semantic search query using examples. The query result provides the top-10 similar clinical cases for use as a prompt thought.

We randomly selected six clinical cases with less than 200 words from the test dataset that contain 1 or more ICD codes or text references. 
The breakdown of the CodiEsp samples along with their ICD codes, text references and word count for each clinical case are shown in Table~\ref{tab:clinical_case_counts} and Table~\ref{tab:clinical_case_details}.

\begin{table}[ht]
\centering
\caption{Clinical Case Samples (Counts)}
\begin{tabular}{lcccc}
\hline
\makecell{\textbf{Clinical} \\ \textbf{Case}} & \textbf{CodiEsp ArticleID} & \makecell{\textbf{ICD} \\ \textbf{Code}} & \makecell{\textbf{Text} \\ \textbf{Reference}} & \makecell{\textbf{Clinical} \\ \textbf{Note}} \\
\hline
A & S0213-12852003000600002-1 & 2  & 2  & 99  \\
B & S1130-05582017000100031-1 & 1  & 1  & 113 \\
C & S1130-01082008001000008-1 & 4  & 5  & 128 \\
D & S1130-01082009000500011-1 & 9  & 9  & 107 \\
E & S1130-01082008000100009-1 & 10 & 11 & 107 \\
F & S1130-01082006001000017-1 & 9  & 9  & 90  \\
\hline
\end{tabular}
\label{tab:clinical_case_counts}
\end{table}

\begin{table}[!htbp]
\centering
\caption{Clinical Case Samples (Details)}
\begin{tabular}{lcccc}
\hline
\makecell{\textbf{Clinical} \\ \textbf{Case}} & \makecell{\textbf{ICD Code}} & \makecell{\textbf{Text Reference}}  & \makecell{\textbf{Age}}  & \makecell{\textbf{Gender}} \\
\hline
A &  R52, K08.89 & pain, toothache & 56 & female  \\
B &  K08.109 & edentulism & 54 & female \\
C &  \makecell{ E11.9 M79.81 \\ R05 T14.8 } &  \makecell{ abdominal wall \\ hematoma, cough, DM \\  ID, hematoma, right \\ anterior rectal hematoma } & 61 & male \\
D &  \makecell{ B00.9 B99.9 \\ C15.9 F17.210 \\ K22.2 R13.10 \\ R50.9 R60.9 \\ R63.4 } & \makecell{ dysphagia, edema, \\ esophageal carcinoma, \\ fever, herpes simplex, \\ infection, partial narrowing \\ of the esophageal \\ lumen, smoker, weightloss } & 67 & male \\
E &  \makecell{ B99.9 C81.90 \\ G93.40 K72.00 \\ K72.90 K92.1 \\ K92.2 R58 \\ R69 R74.0 } & \makecell{ acute liver failure, \\ digestive bleeding, disease, \\ elevated transaminases, \\ encephalopathy, hemorrhage, \\ hepatic encephalopathy, \\ Hodgkin's disease, infection, \\ liver failure, manes } & 16 & male \\
F &  \makecell{ I10 K42.9 \\ K56.60 K92.2 \\ N80.5 R10.32 \\ R10.814 \\ R19.8 R58 }  & \makecell{ AHT, colon endometriosis, \\ colon hemorrhage, \\ hemorrhage, pain in FII, \\ pain on deep palpation \\ in FII, rectal urgency, \\ sigma level stricture \\ imaging, umbilical hernia }  & 42 & female \\
\hline
\end{tabular}
\label{tab:clinical_case_details}
\end{table}

\subsection{Knowledge Graph \& Clinical Cases}

We utilized the SNOMED CT to ICD-10-CM map ~\citep{nlm_snomedct_icd10cm} as the foundation for constructing our KG prompt.  To enrich this mapping, we curated data from the SNOMED CT OWL ontology reference set  ~\citep{nlm_snomedct}, which provides the OWL expressions and the accompanying description file ~\citep{nlm_snomedct}, which provides human-readable terms that convey the intended meaning of each concept.  By integrating SNOMED CT concepts with their corresponding ICD-10-CM codes, OWL expressions, and descriptive labels, we established clinically meaningful links between the ICD codes, ontological definitions, and semantic representations. 

\begin{figure}[ht]
\centering
\includegraphics[width=\linewidth]{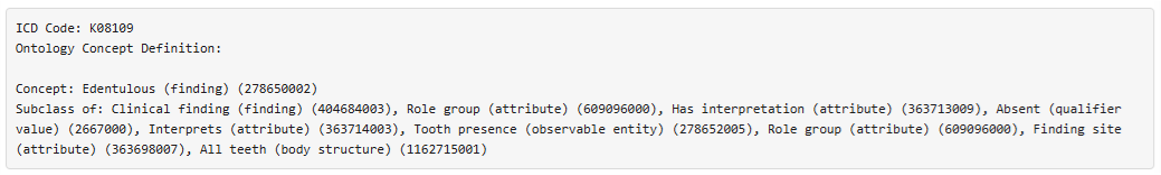}
\caption{An example of the OWL expression prompt for CoT KG prompting.}
\label{fig:fig2}
\end{figure}

Figure~\ref{fig:fig2} illustrates the OWL expression derived from SNOMED CT and corresponds to the ICD-10 code K08.109 (Complete loss of teeth, unspecified cause) within Clinical Case B. It is used to formalize the concept Edentulous (finding) (278650002) using a combination of ontological relationships that capture its clinical semantics. Specifically, the expression indicates the absence (Absent [2667000]) of the observable entity Tooth presence (278652005), located at the finding site All teeth (1162715001). These attributes are grouped under the Role group (609096000), reflecting a normalized structure for reasoning over clinical findings. 

The OWL axioms capture logical relationships and attributes among SNOMED CT concepts, enabling more advanced, ontology-driven reasoning. This alignment results in a structured and semantically rich prompt for GPT-4, facilitating the generation of a clinically accurate and logically coherent knowledge graph that connects diagnoses, procedures, and broader biomedical relationships. From the listing of ICD-10 codes within our curated data, we were able to match the ICD Code: K08.109 (Complete loss of teeth, unspecified cause, unspecified class) to the only ICD code associated to Clinical Case B in Table~\ref{tab:clinical_case_details}.

\subsection{Prompt Format}
\begin{figure}[ht]
\centering
\includegraphics[width=\linewidth]{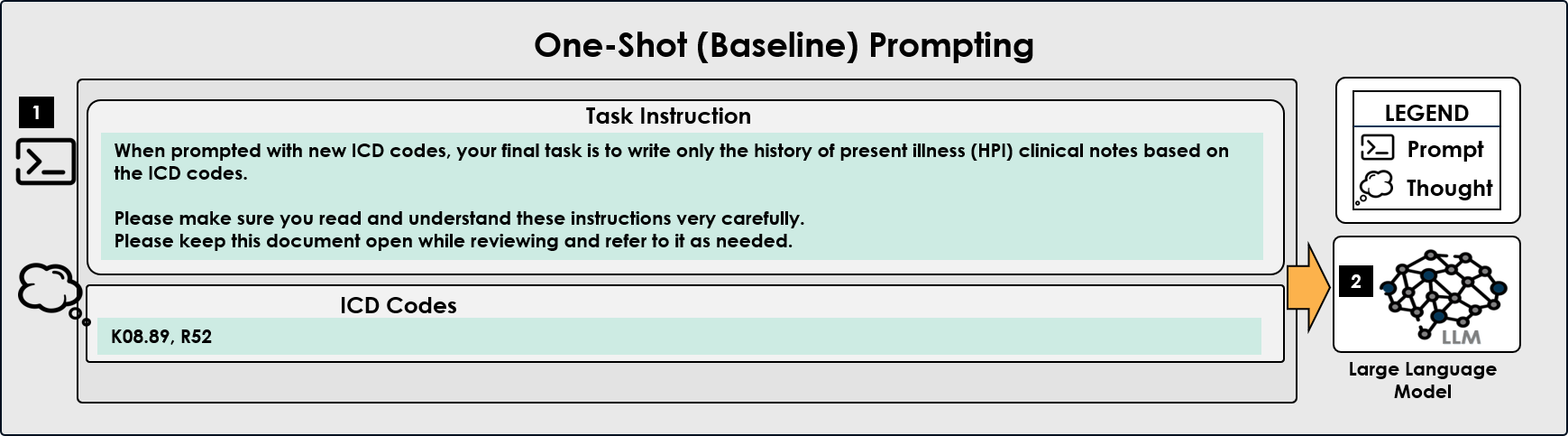}
\caption{Baseline (One-Shot) Prompt.}
\label{fig:fig3}
\end{figure}

\begin{figure}[ht]
\centering
\includegraphics[width=\linewidth]{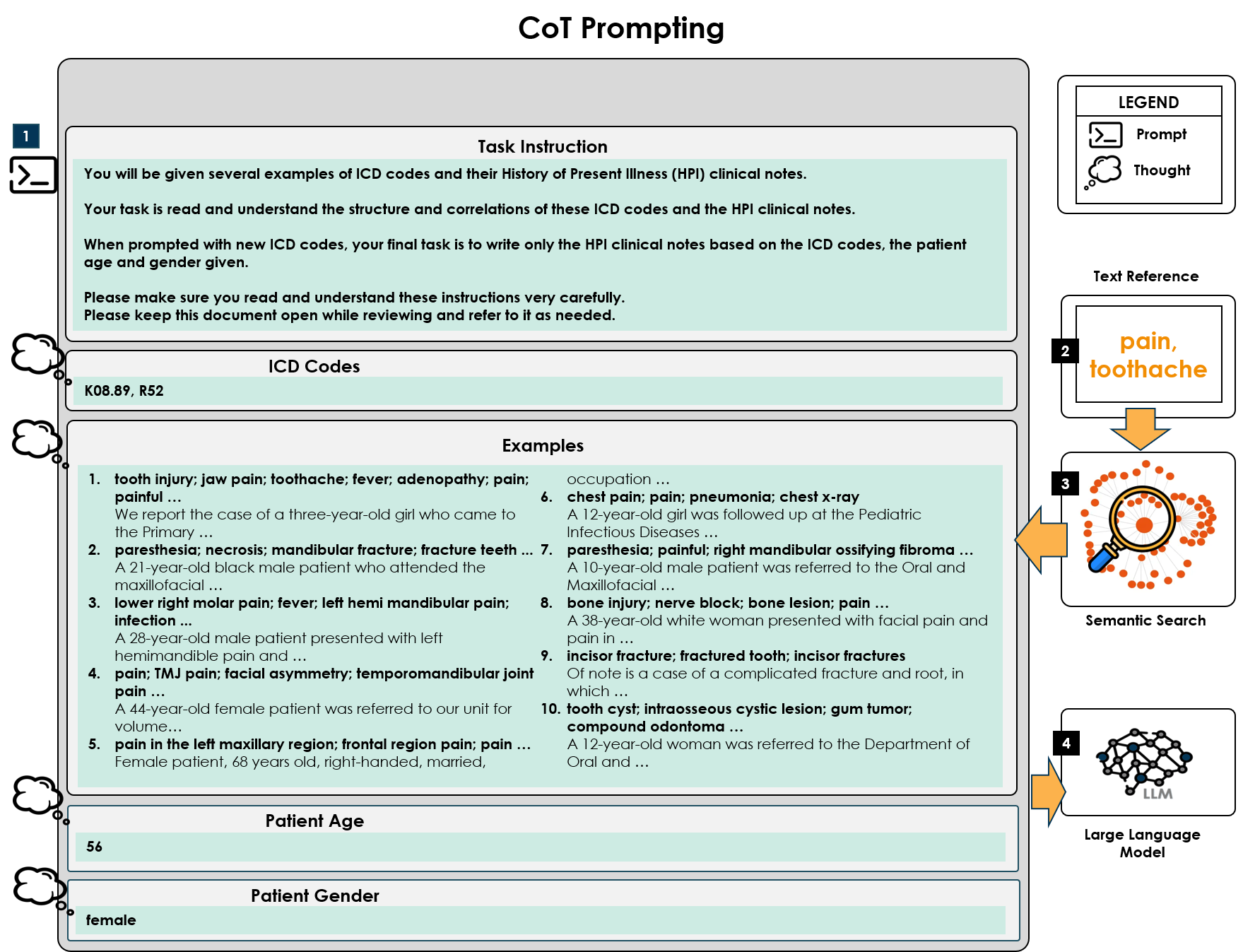}
\caption{CoT Prompt (leveraging the ICD Code semantic search query).}
\label{fig:fig4}
\end{figure}

Our standard prompt, referred to as the baseline, is formatted as task instructions with one-shot prompt to generate the HPI clinical notes based on the given ICD codes shown in Figure~\ref{fig:fig3}.  Our CoT semantic search (CoT prompting) is formatted as instructions to guide the output of language model by controlling its generated text shown in Figure~\ref{fig:fig4}.  
In the CoT instruction prompt, each experiment contains the ground-truth clinical case’s ICD codes along with basic patient information.  

\begin{figure}[ht]
\centering
\includegraphics[width=\linewidth]{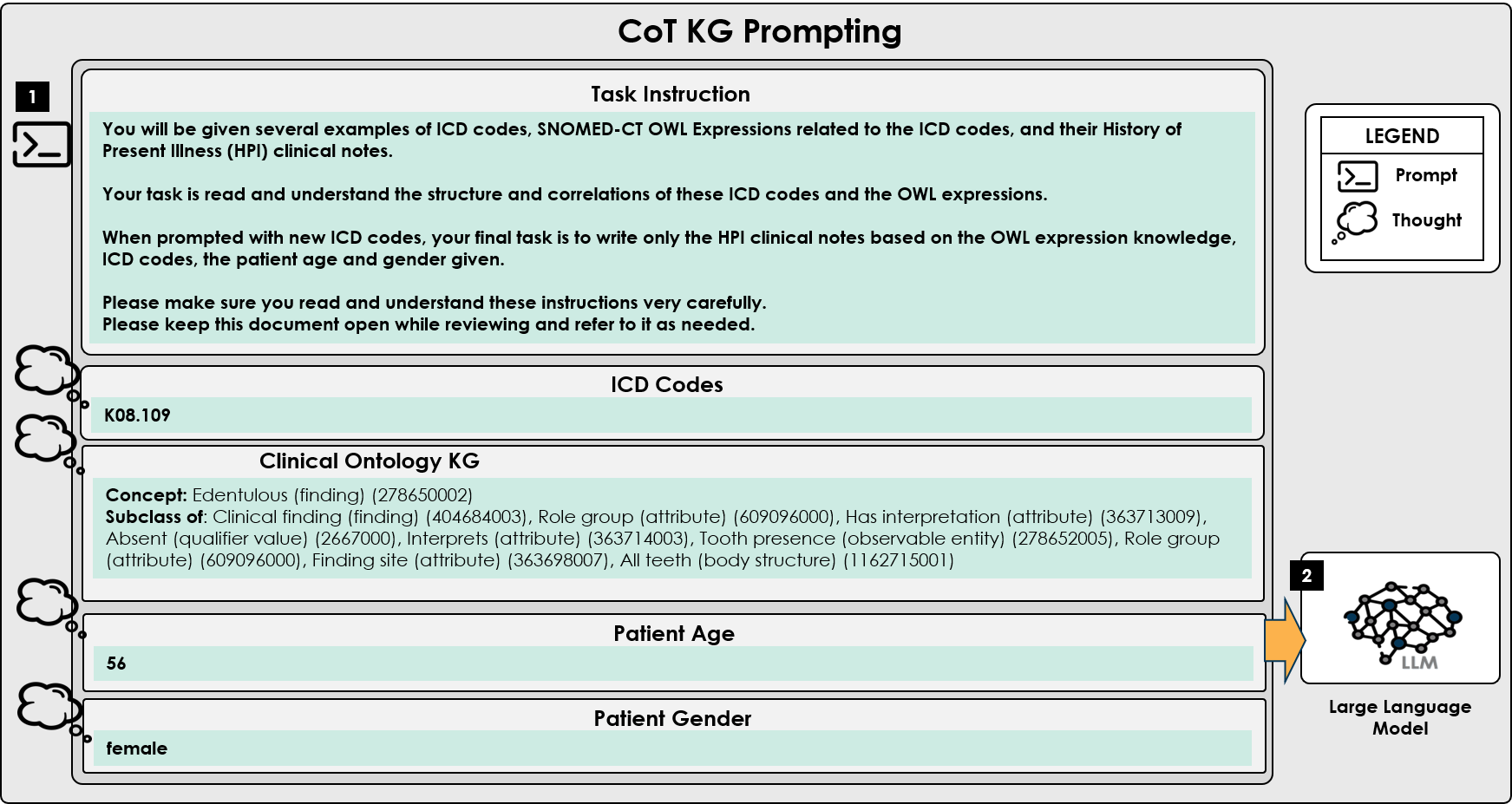}
\caption{CoT Prompt (leveraging the knowledge graph).}
\label{fig:fig5}
\end{figure}

\begin{figure}[ht]
\centering
\includegraphics[width=\linewidth]{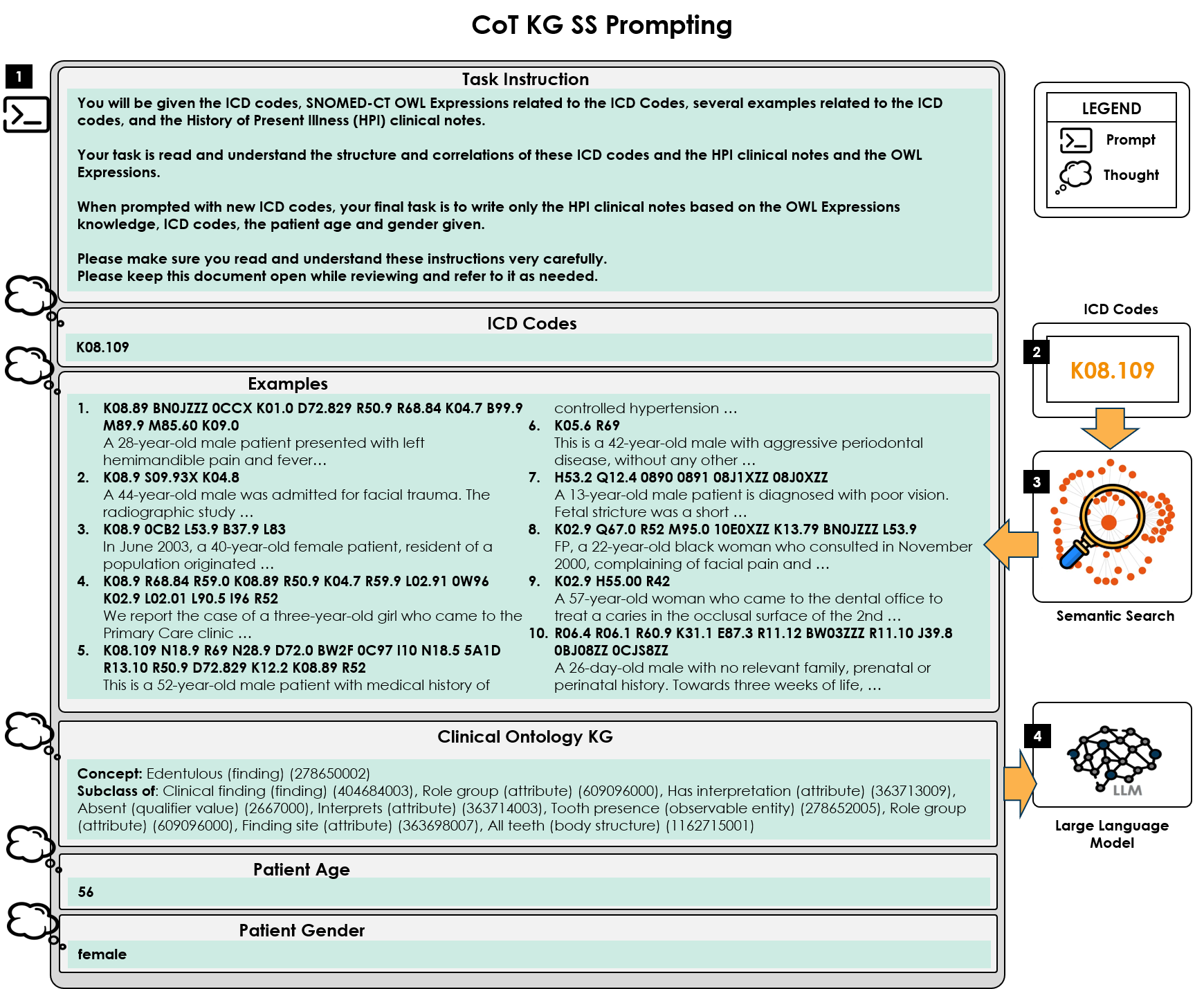}
\caption{CoT Prompt (leveraging the semantic search query and knowledge graph).}
\label{fig:fig6}
\end{figure}

In addition, the top-10 similar clinical cases are provided by semantic search query (based on the ground-truth ICD codes or text references), which uses contextualized word embeddings and the cosine similarity function to find related clinical cases, are provided as prompts.  
These inputs act as rationales, enabling the LLM to learn and generate the intended clinical notes based on the provided ICD codes.  

Our CoT prompting takes inputs, such as: task instruction, ICD codes for the diagnosis and/or procedure, examples of similar clinical cases using the semantic search (ICD codes or text references) query, and Basic patient information (age and gender).
Both CoT KG (Figure~\ref{fig:fig5}) and CoT Semantic Search KG (Figure~\ref{fig:fig6}) prompting techniques utilize the SNOMED-CT Knowledge Graph derived from OWL expressions, as illustrated in Figure~\ref{fig:fig2}. The key difference between the two lies in the use of semantic search: CoT KG does not incorporate semantic search, whereas CoT SS KG does.  

\subsection{Metrics}
Cosine distance is the complement of cosine similarity, which measures the angular difference between two vector representations in a multi-dimensional space. 
It is a mathematical function that quantifies the degree of dissimilarity between two vectors based on their orientation rather than their magnitude. 
The cosine distance formula is defined as:
\begin{equation}
\text{Cosine Distance} = 1 - \frac{\mathbf{A} \cdot \mathbf{B}}{\|\mathbf{A}\| \, \|\mathbf{B}\|} 
= 1 - \frac{\sum_{i=1}^{n} A_i B_i}{\sqrt{\sum_{i=1}^{n} A_i^2} \, \sqrt{\sum_{i=1}^{n} B_i^2}}
\end{equation}
\noindent where, $\mathbf{A} \cdot \mathbf{B}$ is the dot product of the sentence vectors $\mathbf{A}$ and $\mathbf{B}$, $\|\mathbf{A}\|$ and $\|\mathbf{B}\|$ are the magnitudes (norms) of the vectors, and the result gives a measure of the angular distance between the vectors.

Transformer-based models, such as Bidirectional Encoder Representations from Transformers (BERT) ~\citep{devlin2019bert}, capture both syntactic and semantic relationships between words by generating contextualized word embeddings.  To assess the similarity between machine-generated and ground-truth documents, we use BERT. Both documents are processed through the bert-large-cased model to obtain embeddings, which are then used to calculate sentence similarities. The two steps are the following.
\begin{enumerate}
    \item Using the special “classification” [CLS] token of each sentence.  The [CLS] token in BERT serves as a holistic representation of the input sequence. The output of [CLS] is inferred by all other words in this sentence.  This implies that the [CLS] contains all information in other words, which makes [CLS] a representation for sentence-level classification.
    \item Calculating the MEAN of the sentence embeddings provides a way to quantify the overall cosine distance between the sentences based on semantic meaning. 
\end{enumerate}
	
\subsection{Experiments} 
For our experiments, we use OpenAI’s GPT-4 (\texttt{gpt-4}) model for all experiments, with the following parameters: seed (123), temperature (0), top\_p (0.000001), frequency\_penalty (0), and presence\_penalty (0). 
We establish a baseline for our results using the standard one-shot prompt and compare it against the results from our CoT prompts, which utilize a semantic search query based on the provided ICD codes or text references from ground-truth clinical case samples. 
Basic patient information is provided as supplementary prompts. 
Our semantic search query introduces an extra prompt, which includes the top-10 most similar clinical cases based on the given ICD codes or text references.

For each clinical case sample, we collect results from 100 API calls to GPT-4, with each call treated as an independent interaction. 
This ensures there is no memory or history from previous interactions, making each response independent. 
The clinical case’s top-10 relatedness scores from the semantic search query are presented in Table~\ref{tab:relatedness_scores}. 
These scores are calculated using cosine distance, which evaluates spatial proximity to identify the top-10 most similar clinical cases based on the provided ICD codes or text references.

\begin{table}[ht]
\centering
\caption{Semantic Search Query (Top-10 Relatedness Scores)}
\begin{tabular}{l p{5.5cm} p{5.5cm}}
\hline
\makecell{\textbf{Case\#}} & \textbf{ICD Code Relatedness} & \textbf{Text Reference Relatedness} \\
\hline
A & 0.762, 0.754, 0.729, 0.724, 0.718, 0.715, 0.708, 0.695, 0.687, 0.683 & 0.563, 0.483, 0.478, 0.469, 0.462, 0.458, 0.433, 0.431, 0.429, 0.417 \\
\hline
B & 0.720, 0.717, 0.711, 0.701, 0.677, 0.672, 0.653, 0.644, 0.635, 0.630 & 0.469, 0.434, 0.411, 0.387, 0.380, 0.361, 0.359, 0.338, 0.336, 0.334 \\
\hline
C & 0.812, 0.780, 0.775, 0.768, 0.768, 0.760, 0.759, 0.754, 0.754, 0.751 & 0.601, 0.553, 0.545, 0.522, 0.521, 0.520, 0.520, 0.517, 0.512, 0.511 \\
\hline
D & 0.803, 0.802, 0.798, 0.796, 0.787, 0.783, 0.783, 0.782, 0.782, 0.782 & 0.630, 0.613, 0.606, 0.568, 0.565, 0.551, 0.547, 0.537, 0.524, 0.522 \\
\hline
E & 0.846, 0.807, 0.805, 0.795, 0.786, 0.774, 0.770, 0.769, 0.767, 0.757 & 0.637, 0.632, 0.623, 0.613, 0.610, 0.609, 0.601, 0.598, 0.595, 0.594 \\
\hline
F & 0.797, 0.775, 0.764, 0.763, 0.759, 0.752, 0.744, 0.741, 0.739, 0.739 & 0.654, 0.646, 0.645, 0.629, 0.627, 0.625, 0.624, 0.617, 0.615, 0.610 \\
\hline
\end{tabular}
\label{tab:relatedness_scores}
\end{table}

\section{Results and Discussions} 
We evaluate our prompting technique using cosine distance as the primary metric, comparing the generated text against the ground truth.  
We utilized Kernel Density Estimation (KDE) plots to visualize the distribution of our results and support our discussion, Box plots to show the comparison of our CoT prompting against the various CoT KG approach, and Uniform Manifold Approximation and Projection (UMAP) ~\citep{mcinnes2018umap} plots to show a visual comparison of noun distributions between Ground Truth and various generation methods.

\begin{figure}[ht]
\centering
\includegraphics[width=\linewidth]{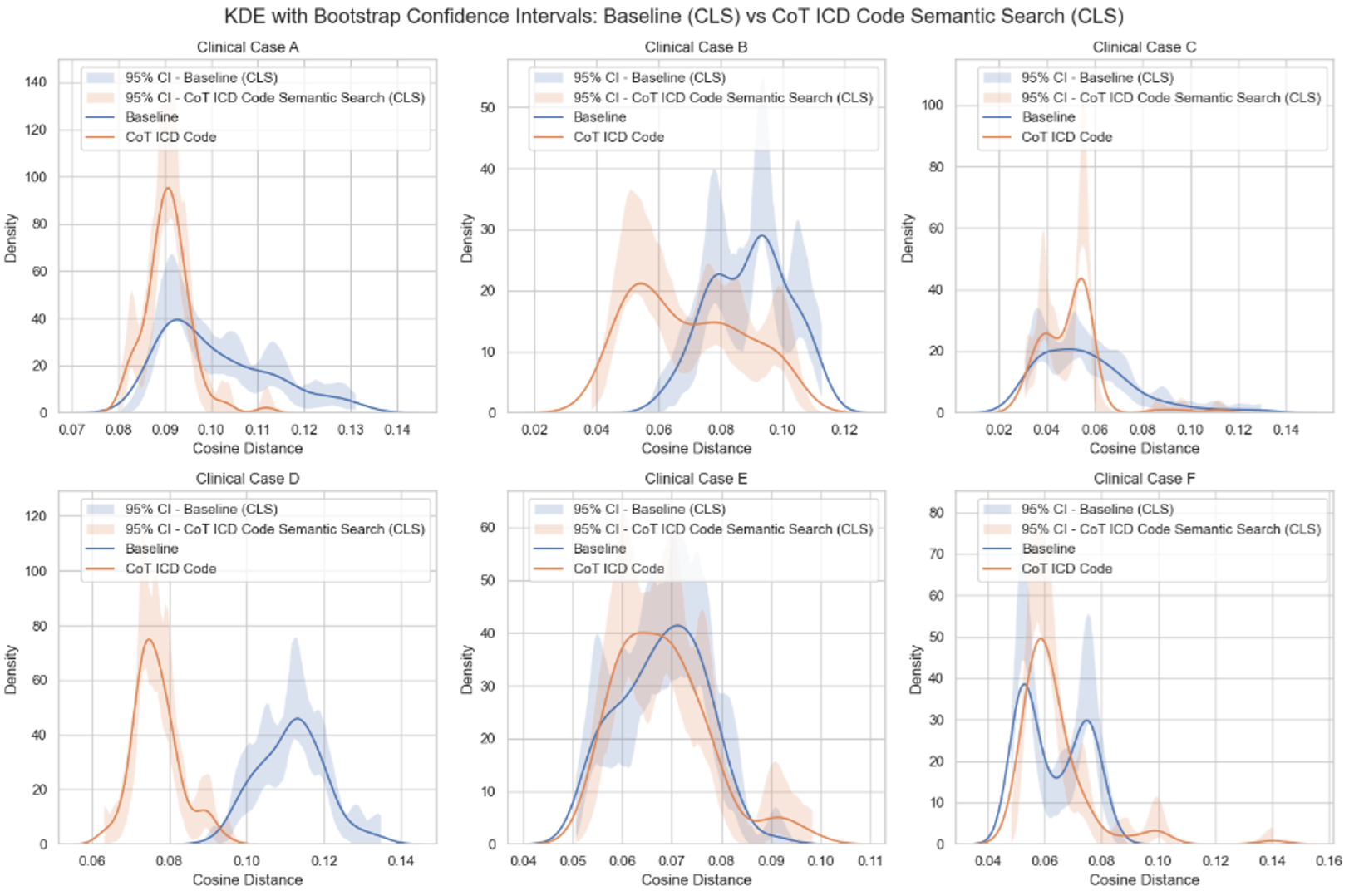}
\caption{Illustration of the six clinical cases using KDE with BCIs to compare the Baseline and CoT ICD code semantic search, based on cosine distance score of sentence-level [CLS].}
\label{fig:fig7}
\end{figure}

\begin{figure}[h]
\centering
\includegraphics[width=\linewidth]{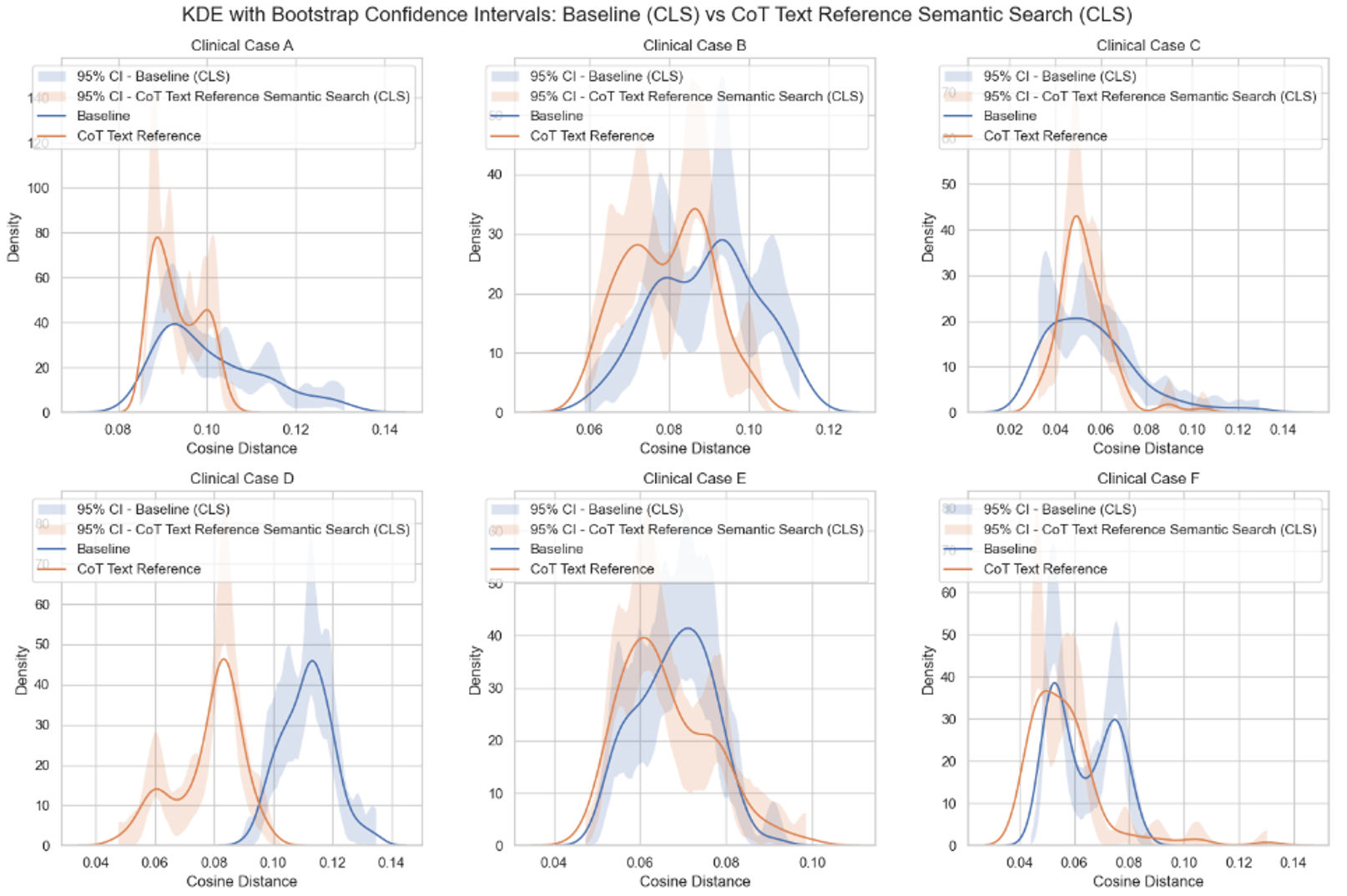}
\caption{Illustration of the six clinical cases using KDE with BCIs to compare the Baseline and CoT text reference semantic search, based on cosine distance score of sentence-level [CLS].}
\label{fig:fig8}
\end{figure}

\begin{figure}[ht]
\centering
\includegraphics[width=\linewidth]{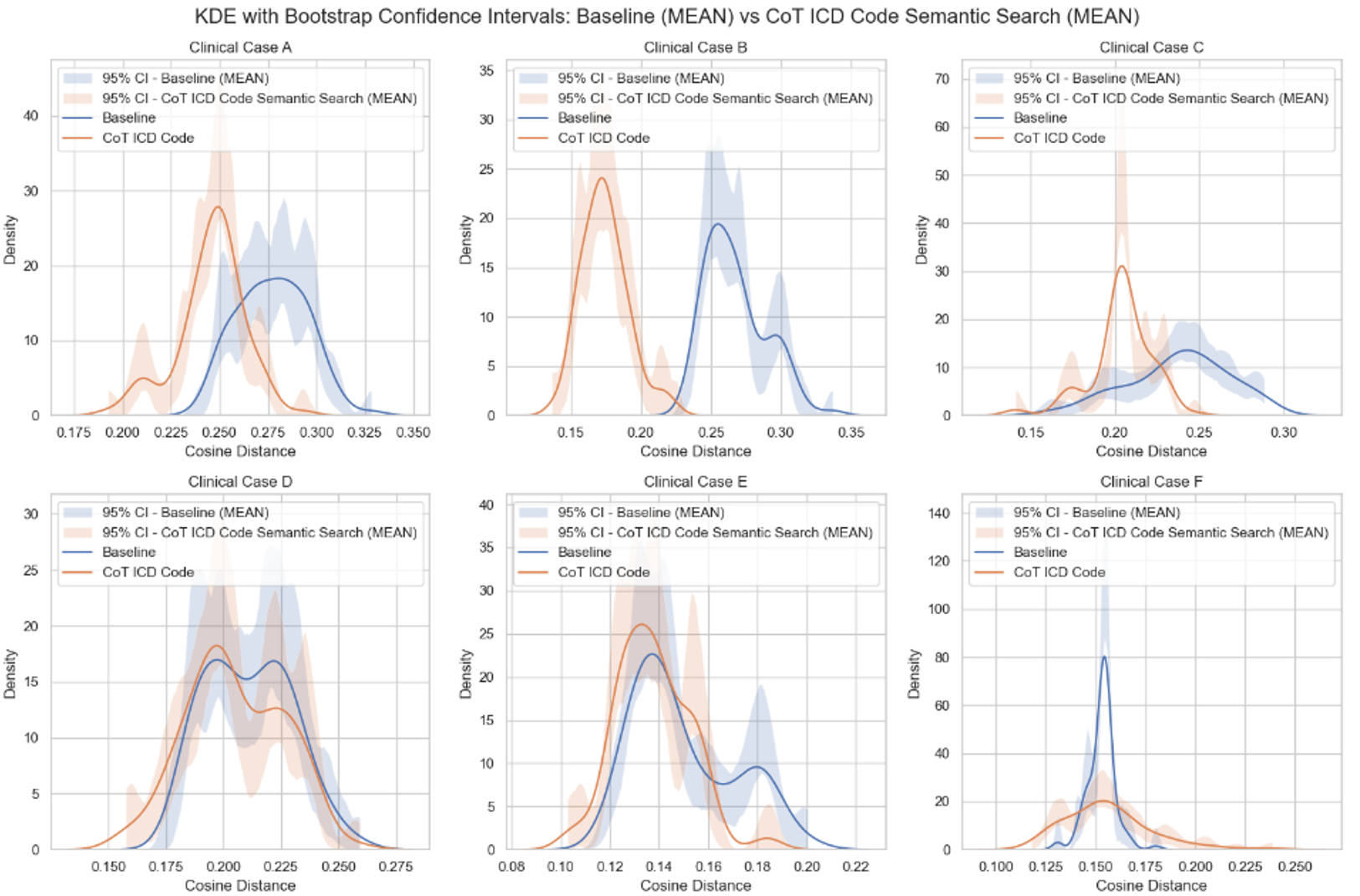}
\caption{Illustration of the six clinical cases using KDE with BCIs to compare the Baseline and CoT ICD code semantic search, based on cosine distance score of sentence-level MEAN.}
\label{fig:fig9}
\end{figure}

\begin{figure}[ht]
\centering
\includegraphics[width=\linewidth]{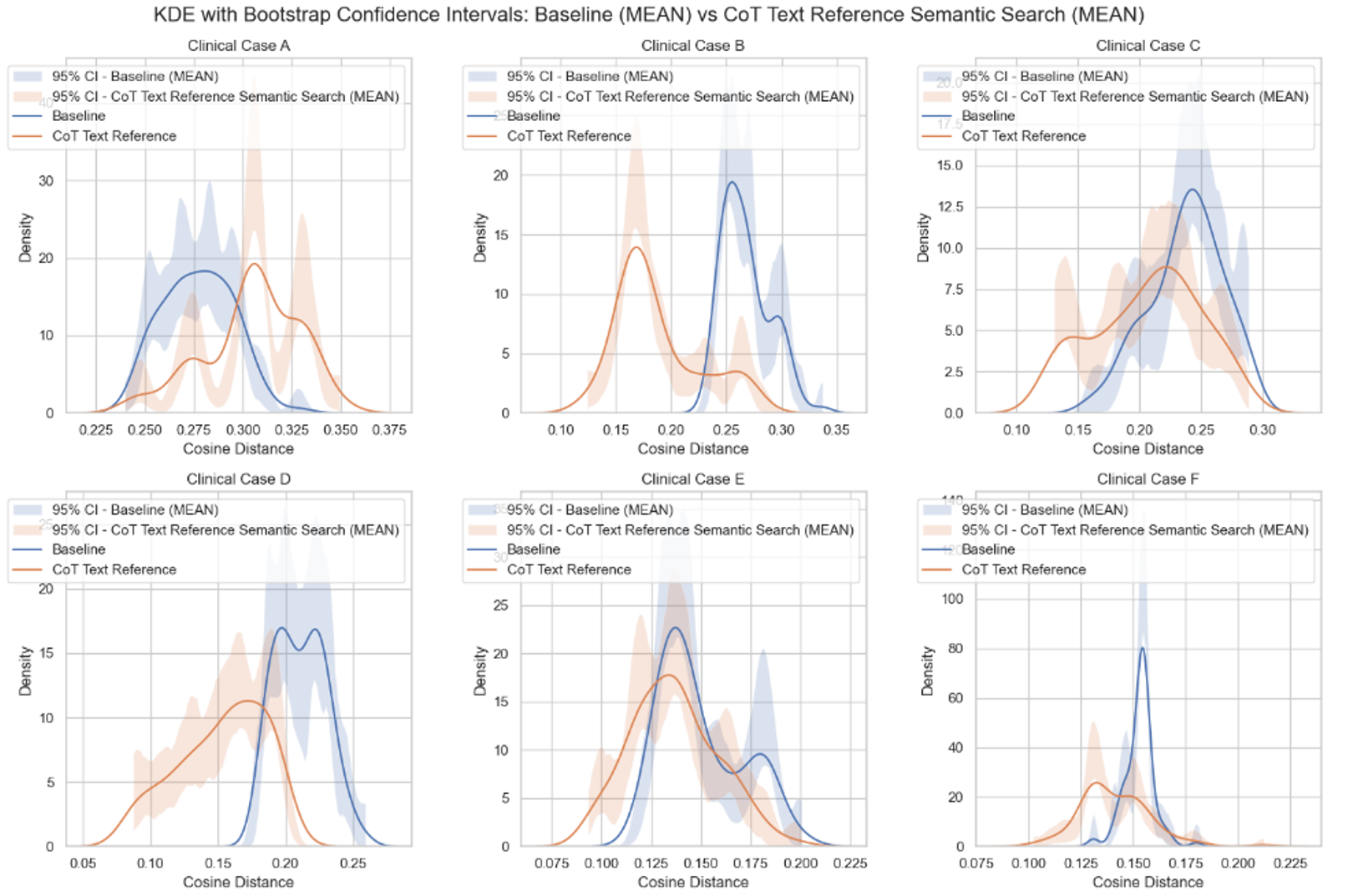}
\caption{Illustration of the six clinical cases using KDE with BCIs to compare the baseline and CoT text reference semantic search, based on cosine distance score of sentence-level MEAN.}
\label{fig:fig10}
\end{figure}

\subsection{KDE Results and Discussions}

We present the results using KDE plots, which include Bootstrap Confidence Intervals (BCIs) for both sentence-level [CLS] and mean scores. BCIs were used to estimate the uncertainty of results by resampling the data with replacement, allowing for robust, distribution-free confidence estimates. This provides a more reliable measure of variability, especially when parametric assumptions may not hold. KDE plots offer a smooth, continuous representation of the underlying data distribution, allowing for clearer identification of patterns, peaks, and differences between groups. Unlike histograms, which can be sensitive to bin size and may obscure subtle features in the data, KDE provides a more refined view that enhances interpretability. This visualization enables a direct comparison between our CoT prompting approach and the baseline one-shot prompt. The distributional analysis shows that CoT prompting, enhanced with semantic search queries such as ICD codes and relevant text references, consistently improves the model's ability to capture underlying clinical reasoning. This includes better representation of structured information from ICD codes and contextual patient details, leading to superior performance compared to the baseline.

The KDE with BCI plots (Figures~\ref{fig:fig7}-~\ref{fig:fig10}) reveal a leftward shift in the peaks for the CoT semantic search prompting technique, indicating that its distribution has lower values compared to the baseline prompts.  
This shift highlights notable differences in semantic alignment with ground-truth clinical cases. 
The inclusion of BCIs provides statistical validation, enhancing the robustness and interpretability of these findings.  
However, as shown in Figure ~\ref{fig:fig10} , we observe that in clinical case A, the baseline prompt works better than the CoT text reference prompt. 
This is due to the presence of two text references (pain, toothache). 
In contrast, in Figure ~\ref{fig:fig8} , the two ICD codes (K08.89, R52) clinical case A perform better than the baseline prompt. 

One limitation we observed with CoT prompting is its sensitivity to ambiguous or underspecified clinical terms, such as ``pain” or ``toothache.” 
These terms are semantically broad and often lack contextual qualifiers like location, severity, or duration, which are essential for generating accurate and clinically useful notes. 
For example, the ICD-10 code R52 (Pain, unspecified) is generally considered too broad for clinical documentation because it fails to capture the underlying source or etiology of the pain and may even raise compliance concerns. 
In contrast, when the record specifies dental pain or pain due to a tooth disorder, K08.89 (Other specified disorders of teeth and supporting structures) is the more clinically meaningful choice, making R52 redundant unless no other ICD code provides greater specificity. 
In such edge cases, CoT prompting can either propagate the ambiguity of the input or reflect limitations in the annotated data, defaulting to overly broad codes when faced with vague terms.


\subsection{CoT KG Results and Discussion}

To summarize and compare the performance across the baseline and CoT prompting methods, we use box plots to visualize the results collected from 250 API calls to GPT-4 for our CoT KG setup. These plots provide a concise representation of the output distribution, highlighting key statistical features such as the median, interquartile range (IQR), and potential outliers, thereby facilitating an informative comparison of variability and central tendency across prompting methods.

\begin{figure}[ht]
\centering
\includegraphics[width=\linewidth]{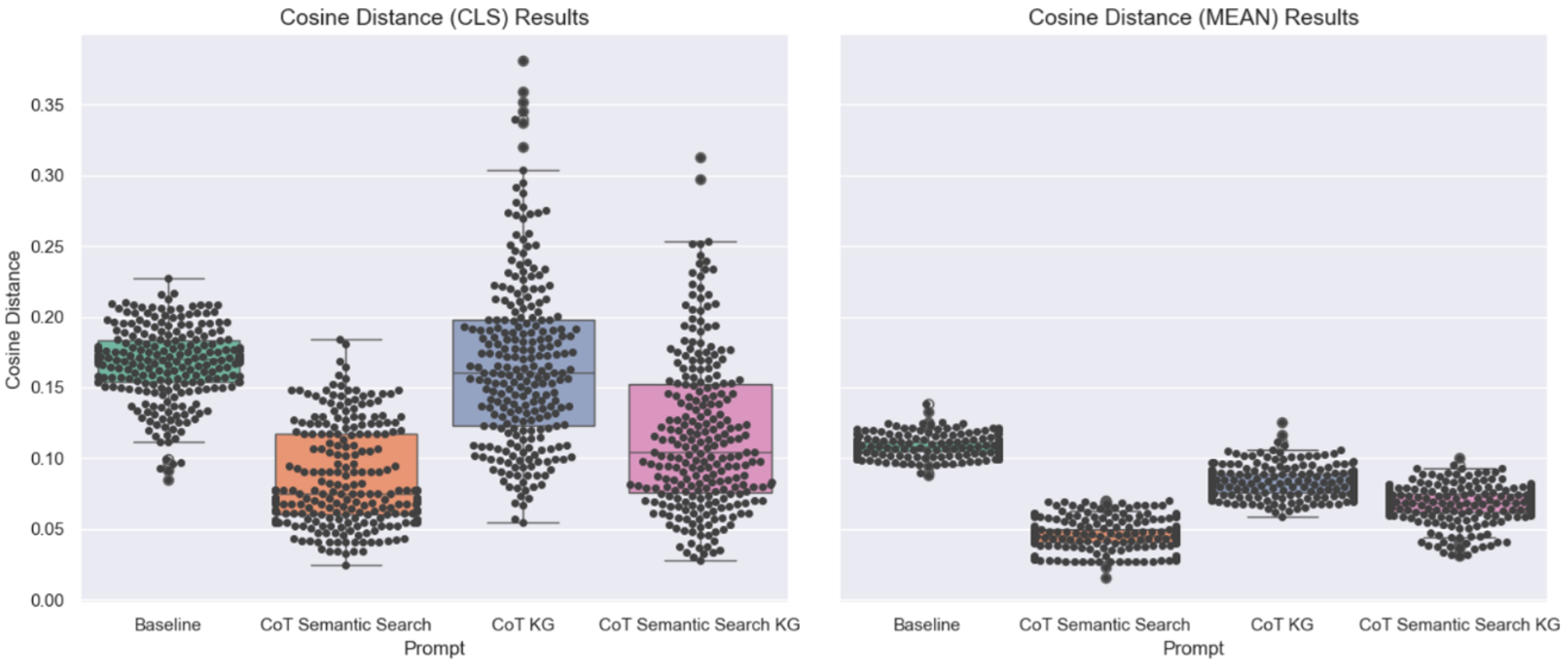}
\caption{Illustration of the KG results against the Baseline and CoT prompting for sentence-level CLS and MEAN.}
\label{fig:fig11}
\end{figure}

In Figure~\ref{fig:fig11}, the results of [CLS] (left) show that CoT prompting yields the lowest cosine distances, while CoT KG leads to the highest distances with the greatest variance. In the [MEAN] results (right), CoT prompting achieves the lowest distances, and all prompts exhibit tighter distributions and improved semantic similarity.
These findings suggest that, although the ontology-based KG introduces greater output diversity, it does not significantly enhance the precision or semantic accuracy of the generated medical notes.

\subsection{Part-of-Speech (PoS) Tagging (Nouns and PROPN) Results and Discussion}

To better understand why CoT prompting outperforms both CoT-KG approaches in aligning ground-truth and generated outputs, we applied UMAP ~\citep{mcinnes2018umap} along with the transformer-based model SciBERT ~\citep{beltagy2019scibert} to visualize noun embeddings across different prompting strategies. For each of five sampled instances, we generated a separate plot comparing all four prompting techniques. Each technique's visualization was based on a single randomly selected run from its 250 generations.  The UMAP dimensionality reduction technique allows us to project high-dimensional token-level embeddings into a 2D-space, facilitating direct visual comparison of distributions. We focused specifically on PoS tagging (NOUN and PROPN), as they often carry the most critical common and proper noun meaning within the tokens.  The UMAP parameters: n\_neighbors (15), min\_dist (0.1), and random\_state (1) were used to emphasize local structure and ensuring reproducibility.

\begin{figure}[ht]
\centering
\includegraphics[width=\linewidth]{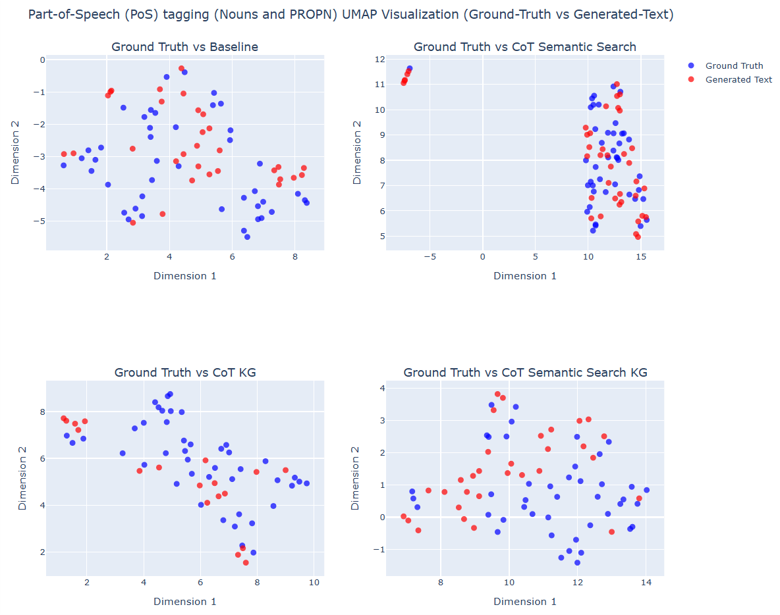}
\caption{UMAP Token Projection for Sample Instance \#1.}
\label{fig:fig12}
\end{figure}

\begin{figure}[ht]
\centering
\includegraphics[width=\linewidth]{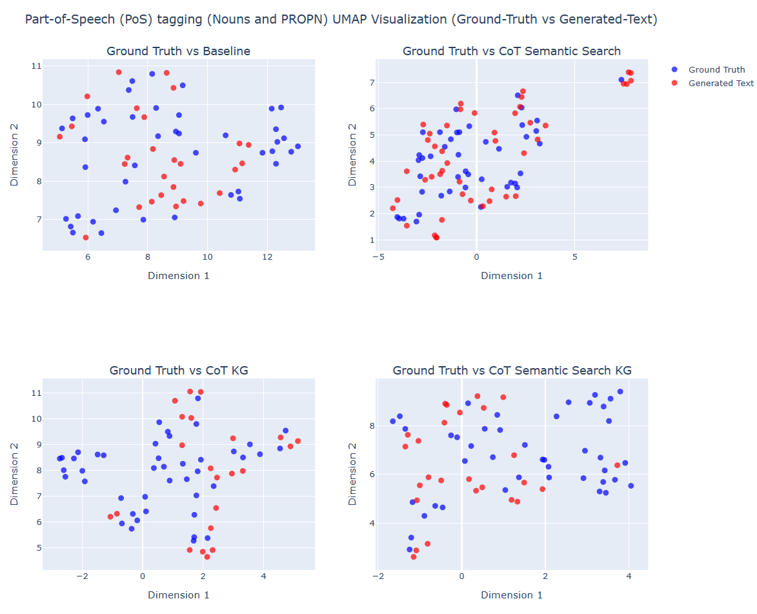}
\caption{UMAP Token Projection for Sample Instance \#2.}
\label{fig:fig13}
\end{figure}

\begin{figure}[ht]
\centering
\includegraphics[width=\linewidth]{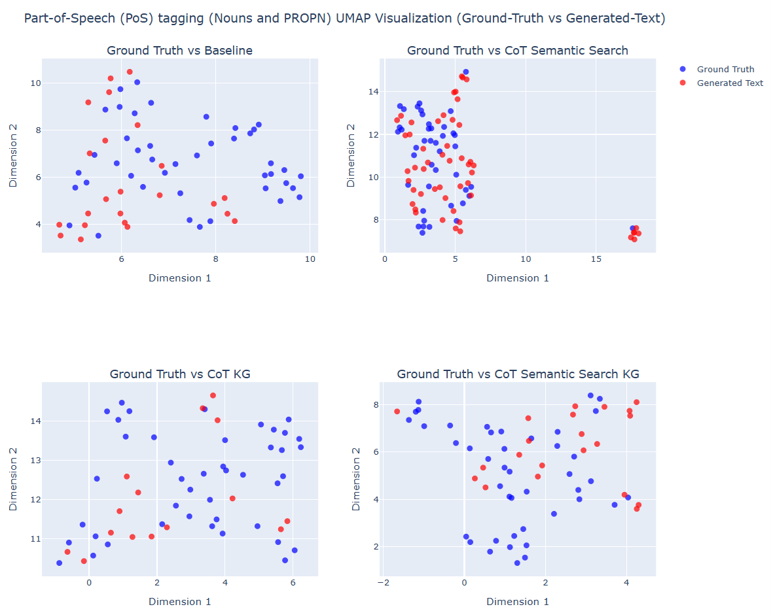}
\caption{UMAP Token Projection for Sample Instance \#3.}
\label{fig:fig14}
\end{figure}

\begin{figure}[ht]
\centering
\includegraphics[width=\linewidth]{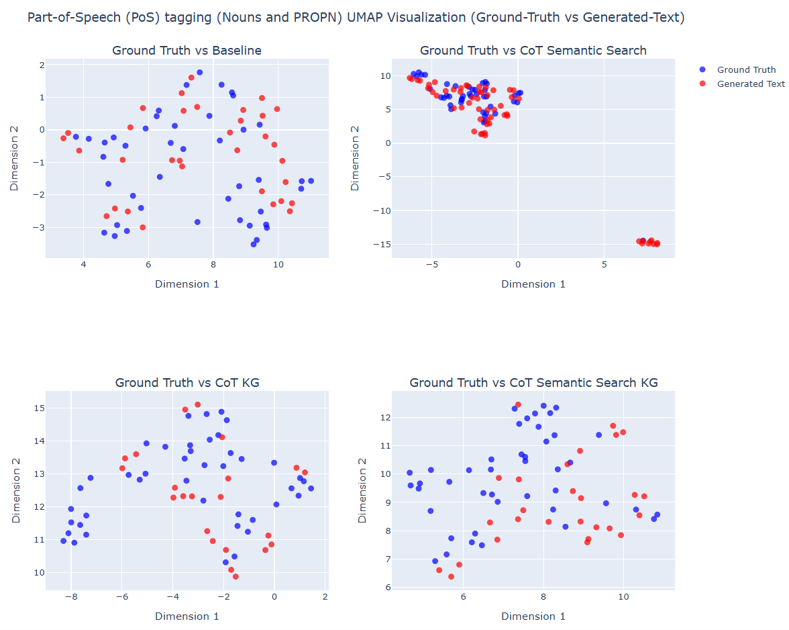}
\caption{UMAP Token Projection for Sample Instance \#4.}
\label{fig:fig15}
\end{figure}

\begin{figure}[ht]
\centering
\includegraphics[width=\linewidth]{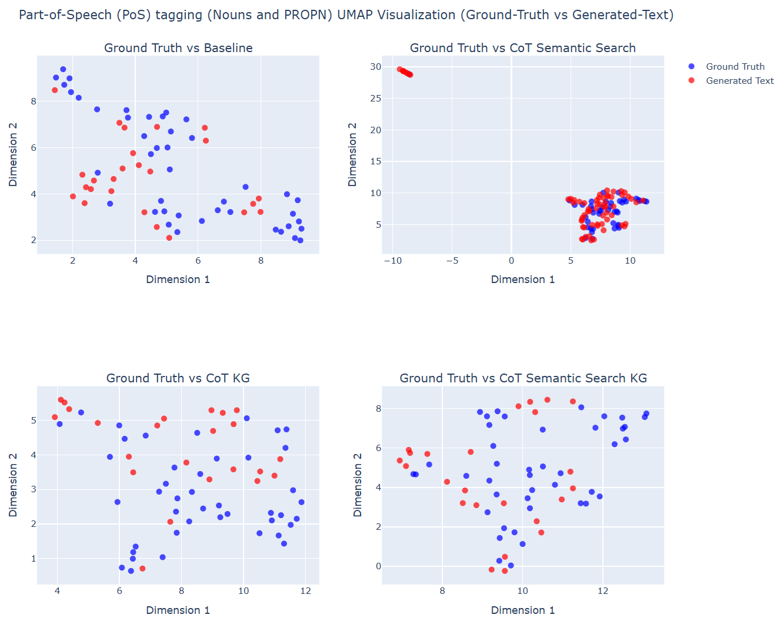}
\caption{UMAP Token Projection for Sample Instance \#5.}
\label{fig:fig16}
\end{figure}

In Figures~\ref{fig:fig12}~\ref{fig:fig13}~\ref{fig:fig14}~\ref{fig:fig15}~\ref{fig:fig16}, the illustrations shows the UMAP distribution of NOUN and PROPN (proper noun) tokens from ground-truth and machine-generated outputs, where each subplot shows one representative result for a different prompt technique.  In each figure, the Baseline output displays broader dispersion and limited overlap between generated and ground-truth tokens, indicating greater semantic divergence. In contrast, the CoT-based prompting techniques demonstrate tighter clustering and increased overlap, suggesting improved semantic alignment of the knowledge. Like the Baseline, the CoT KG plots shows moderate overlap but exhibits some spatial drift between token clusters, which may reflect subtle content hallucinations shifts.  The CoT prompting technique appears to leverage the clinical knowledge from the semantic search examples.

\subsection{Clinical Case Example}

\begin{figure}[ht]
\centering
\includegraphics[width=\linewidth]{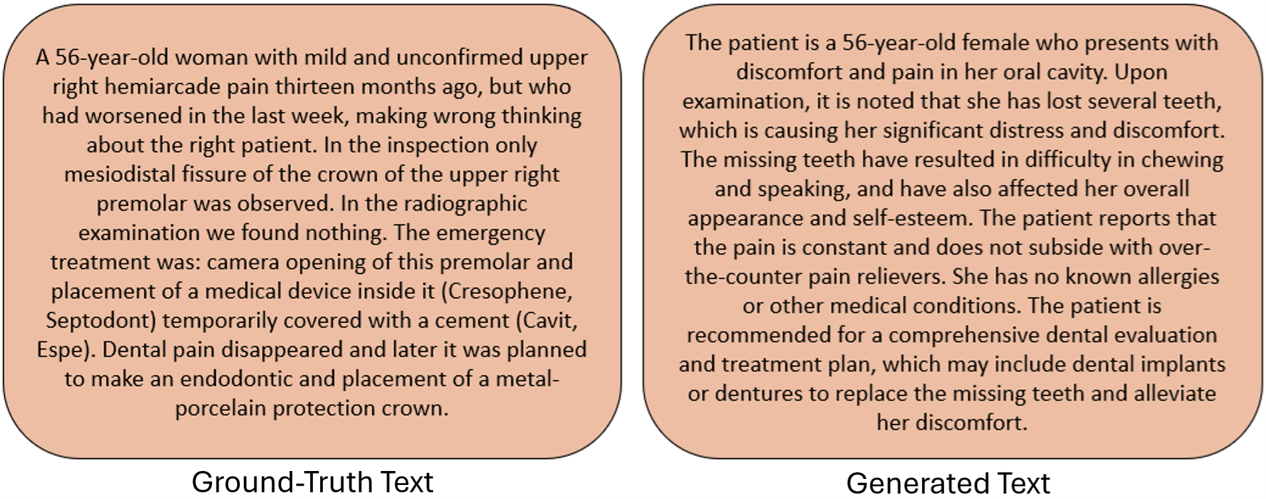}
\caption{A sample from Clinical Case A showing ground-truth and the machine-generated text.}
\label{fig:fig17}
\end{figure}

\begin{figure}[ht]
\centering
\includegraphics[width=\linewidth]{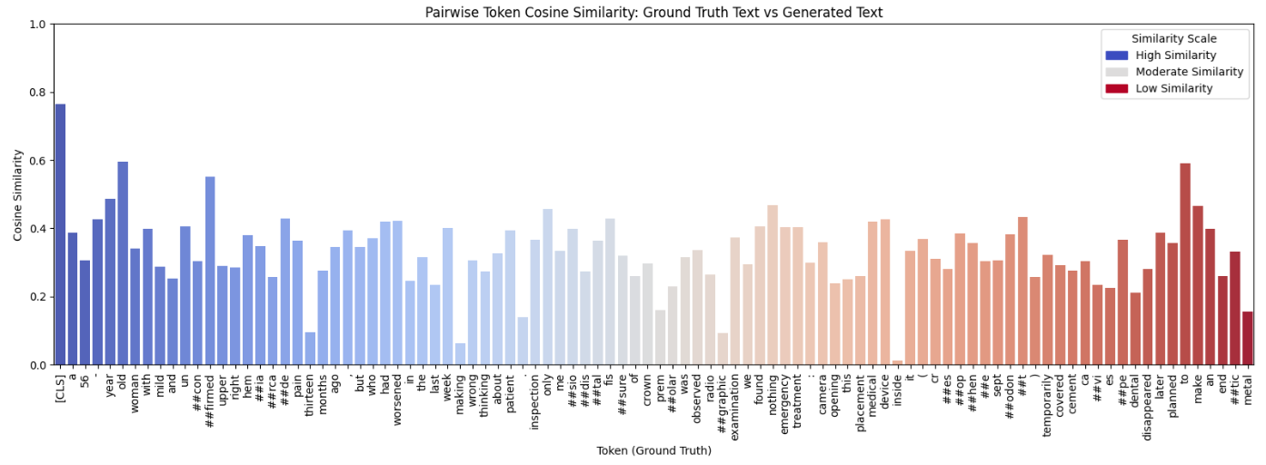}
\caption{Pairwise token-level cosine similarity between ground truth and generated text from Clinical Case A.}
\label{fig:fig18}
\end{figure}

Figure~\ref{fig:fig17} shows that some words are present in the ground-truth but missing from the generated text. 
It also reveals that ICD-10 codes K08.89 (pain) and R52 (toothache) in Table~\ref{tab:clinical_case_details} are linked to terms like ``oral,” ``cavity,” and ``discomfort,” which appear in the generated text, all related to oral health.  
Both texts include ``A 54-year-old female,” confirming alignment in basic patient information. 
Figure~\ref{fig:fig18} presents a pairwise token-level cosine similarity analysis between the ground truth text and generated text, with similarity distinguished by height and color. Each bar represents the cosine similarity of a token from the ground truth (x-axis) with its corresponding token in the generated output. Bar colors indicate similarity levels: blue for high similarity, gray for moderate similarity, and red for low similarity. Higher bars reflect a stronger lexical or semantic alignment between the ground truth and the generated text.  
Several tokens such as ``woman", ``mild", and ``pain" fall within moderate to high similarity (0.3-0.6), suggesting that the model captured clinical terms.  Tokens, such as ``inspection", ``crown", and ``exam" in the (0.2-0.4) range, indicate partial alignment.  
The red bars (right) correspond to lower similarity scores, indicating a greater deviation from the ground-truth text.

\subsection{Results Summary}

Overall, by using visual comparisons, such as overlaying KDE plots for all clinical cases and prompting techniques, we were able to assess the extent of the observed shifts. These plots, along with the cosine distance measurements, confirm that the differences in semantic similarity performance between the techniques are both statistically significant and practically meaningful.
For instance, clinical cases A and B exhibit a significant leftward shift in the peaks (Figures~\ref{fig:fig7} and~\ref{fig:fig10}) for the ICD code semantic search, indicating that their distributions have lower values compared with the baseline. The ICD code prompting technique appears to outperform the Text Reference, likely due to higher relatedness scores for the clinical case examples in Table~\ref{tab:relatedness_scores}.  
From observation, these visual insights combined with the precision offered by the BCIs show that both CoT prompting technique results distributions differ from the baseline standard prompt, which suggest that this technique can guide the generation of clinical notes through instruction prompt using similar clinical cases.

\section{Conclusion and Future Work} 
Our study constructs the HPI clinical notes using CoT prompting with ICD codes, clinical case examples, clinical ontology KG, and basic patient information. 
Experiments were conducted across various clinical cases (clinical cases with 1 ICD code, 2 ICD codes, and several cases with multiple ICD codes), comparing results obtained from a baseline one-shot prompt and two CoT prompting templates.
Through cosine distance analysis, we compared the generated text with ground-truth text, addressing whether LLMs can effectively reason about ICD codes to produce clinical notes using CoT prompting. 

Our analysis concludes that the GPT-4 LLM is capable of reasoning about ICD codes using our CoT semantic search prompting techniques over the baseline one-shot prompt to produce clinical notes.
Also, comparing an EHR to a single ground-truth may not be effective, as different doctors write EHRs differently. 
For this reason, human evaluation, either through expert review, structured validation protocols, or human-in-the-loop feedback, should complement automatic metrics to ensure clinical relevance, factual accuracy, and real-world utility. This perspective is supported by recent work. ~\citep{Tam2024} propose a framework for human evaluation of LLMs in healthcare, highlighting that automatic metrics alone cannot capture critical dimensions such as safety and clinical appropriateness. Similarly, ~\citep{Chiu2024} outline a protocol for human evaluation of generative AI chatbots in clinical consultations, stressing the importance of domain-specific, structured human validation. Designing a human-in-the-loop evaluation phase to complement our current quantitative analyses will be one of our future directions.

Although our results show that the clinical ontology KG can improve over the baseline, it does not further contribute to more precise generation of medical notes when COT with semantic search is incorporated, mainly due to the increased variability of output generated. 
How to effectively incorporate clinical ontology KG into prompting and how to optimally extract and align model outputs with respect to the KG will be the key focus of our future research. The code and materials used for model replication are available at our \href{https://github.com/Moh-najafi14/COT-KG}{GitHub repository} (\url{https://github.com/Moh-najafi14/COT-KG}).

To enhance the prompting techniques, we propose some follow up studies not only in areas of using other instruction prompting techniques, but in these areas as well to enrich LLMs reasoning:
\begin{enumerate}
    \item \textbf{CoT Prompting using Patient’s Past Medical History} \\
The Medical Information Mart for Intensive Care (MIMIC-III)~\citep{johnson2016mimic} offers clinical data on 30-day ICU readmissions, allowing past medical history and admission notes to guide model predictions for future visits.
    \item \textbf{Fine-Tune an LLM to become more biased towards the Physician’s output} \\
LLMs are prone to biases from training data, but they can be fine-tuned for individual physicians using personalized data. This adjustment, achieved through instruction prompting, allows the model to better meet specific needs. Physician notes can be extracted for this fine-tuning from the MIMIC-III dataset.
    \item \textbf{Use RAG in conjunction with CoT prompting} \\
RAG enables retrieval of relevant information, such as patient data from medical databases, to inform the instruction prompting process.  Our semantic search embeddings identify the most relevant documents based on query similarity (e.g., patients with similar ICD codes).
This helps guide LLM text generation, while RAG minimizes ``hallucinations” by feeding relevant facts into the model, improving the accuracy and relevance of clinical note generation.  
    \item \textbf{Better Integrating Disease and Clinical Ontologies as KG Reasoners for Clinical Inference} \\
Incorporating structured and standardized representations of clinical and biomedical knowledge through disease ontologies (e.g., SNOMED CT~\citep{nlm_snomedct_icd10cm, nlm_snomedct}, MONDO~\citep{vasilevsky2022mondo}) has the potential to enhance an LLM’s reasoning by providing a formal framework of relational disease concepts.
We plan to investigate more effective methods for embedding ontology KGs into CoT prompting, such as translating them into natural language descriptions or designing more precise graph-based representations, enabling the LLMs to better infer latent relationships between conditions, symptoms, and comorbidities for clinical note generation.
Furthermore, optimally aligning model outputs with patient-specific details and related conditions in the medical notes could further improve the quality of the generated content.
\end{enumerate}

\section{Acknowledgments}
The project is partially supported by the AI Initiative funded by Old Dominion University. 

\section{Contribution}
Ivan Makohon: Conceptualization, Methodology, Software, Data Curation, Formal Analysis, Writing – Original Draft Preparation, Visualization.
Mohamad Najafi: Software, Writing – Reviewing and Editing.
Jian Wu: Methodology, Writing – Reviewing and Editing, Funding Acquisition.
Mathias Brochhausen: Writing – Reviewing and Editing.
Yaohang Li: Supervision, Project Administration, Conceptualization, Methodology, Writing – Reviewing and Editing, Funding Acquisition.

\section{Conflict of Interest}
None. 

\bibliographystyle{plainnat}
\bibliography{sample}

\end{document}